\documentclass[10pt,twocolumn,letterpaper]{article}

\usepackage{cvpr}
\usepackage{times}
\usepackage{epsfig}
\usepackage{graphicx}
\usepackage{amsmath}
\usepackage{amssymb}
\usepackage{dsfont}
\usepackage{multirow}
\usepackage{svg}

\usepackage{mathtools, nccmath}

\usepackage[normalem]{ulem}
\useunder{\uline}{\ul}{}

\usepackage[pagebackref=true,breaklinks=true,letterpaper=true,colorlinks,bookmarks=false]{hyperref}
\usepackage[capitalise,noabbrev]{cleveref}

\newcommand{\squishitemize}{
\begin{list}{$\bullet$}
	{ \setlength{\itemsep}{0pt}
		\setlength{\parsep}{3pt}
		\setlength{\topsep}{0pt}
		\setlength{\partopsep}{0pt}
		\setlength{\leftmargin}{1.95em}
		\setlength{\labelwidth}{1.5em}
		\setlength{\labelsep}{0.5em} } }
	
\newcounter{Lcount}
\newcommand{\squishlist}{
	\begin{list}{\arabic{Lcount}. }
		{ \usecounter{Lcount}
			\setlength{\itemsep}{0pt}
			\setlength{\parsep}{3pt}
			\setlength{\topsep}{0pt}
			\setlength{\partopsep}{0pt}
			\setlength{\leftmargin}{2em}
			\setlength{\labelwidth}{1.5em}
			\setlength{\labelsep}{0.5em} } }
	
	\newcommand{\squishend}{\end{list}}

\newcommand{\PP}[1]{
	\vspace{2px}
	\noindent{\bf\textsc{#1}}\xspace
}

\newcommand{\model}{\textsc{GLAMOR}\xspace}

\DeclareMathOperator{\sgn}{sgn}
\DeclareMathSymbol{\shortminus}{\mathbin}{AMSa}{"39}

\newenvironment{myequ}{%
	\smallskip\par\centering$\displaystyle}
{$\smallskip\par}

\def\Snospace~{\S{}}

\makeatletter
\newcommand{\crefnames}[3]{%
	\@for\next:=#1\do{%
		\expandafter\crefname\expandafter{\next}{#2}{#3}%
	}%
}
\makeatother

\crefformat{chapter}{\S#2#1#3}
\crefmultiformat{chapter}{\S\S#2#1#3}{and~#2#1#3}{, #2#1#3}{, and~#2#1#3}

\crefformat{section}{\S#2#1#3}
\crefmultiformat{section}{\S\S#2#1#3}{and~#2#1#3}{, #2#1#3}{, and~#2#1#3}

\cvprfinalcopy
\def\cvprPaperID{****} % *** Enter the CVPR Paper ID here

\ifcvprfinal\fi
\begin{document}

%%%%%%%%% TITLE
\title{Looking GLAMORous: Vehicle Re-Id in Heterogeneous Cameras Networks with Global and Local Attention}

\author{Abhijit Suprem\\
Georgia Institute of Technology\\
Atlanta, GA\\
{\tt\small asuprem@gatech.edu}
\and
Calton Pu\\
Georgia Institute of Technology\\
Atlanta, GA\\
{\tt\small calton.pu@cc.gatech.edu}
}

\maketitle
%\thispagestyle{empty}

%%%%%%%%% ABSTRACT
\begin{abstract}
Vehicle re-identification (re-id) is a fundamental problem for modern surveillance camera networks. Existing approaches for vehicle re-id utilize global features and local features for re-id  by combining multiple subnetworks and losses. In this paper, we propose \model, or Global and Local Attention MOdules for Re-id. \model performs global and local feature extraction simultaneously in a unified model to achieve state-of-the-art performance in vehicle re-id across a variety of adversarial conditions and datasets (mAPs 80.34, 76.48, 77.15 on VeRi-776, VRIC, and VeRi-Wild, respectively). \model introduces several contributions: a better backbone construction method that outperforms recent approaches, group and layer normalization to address conflicting loss targets for re-id, a novel global attention module for global feature extraction, and a novel local attention module for self-guided part-based local feature extraction that does not require supervision. Additionally, \model is a compact and fast model that is 10x smaller while delivering 25\% better performance. 
\end{abstract}

%%%%%%%%% BODY TEXT
\section{Introduction}
\label{sec:intro}
Tracking movement of vehicles across multiple cameras in surveillance videos (vehicle re-id) is an important problem in smart transportation and smart cities. Typically re-id systems track vehicles in a frame and evaluate with fixed data sets \cite{RN1}. Vehicles movement is tracked with pixel analysis to find objects with sufficient similarity/proximity \cite{RN32, RN41}. When a vehicle exits a camera, hand-over procedure to the next camera assumes overlapping regions \cite{RN39}.  

%It is a common assumption in machine learning (ML) work that algorithms should be evaluated on fixed and cleaned data sets (with well-known ground truth) for reproducible and comparable accuracy and performance results. This assumption includes a corollary that the gap between the carefully cleaned data sets and noisy real world data is beyond the scope of the study, and therefore it can be left to someone else, for example, through data cleaning methods \cite{RN99}. This scenario applies to offline retrospective studies, or after-the-fact forensics applications. 

\PP{Heterogeneous Camera Networks} In this paper, we focus on real-time vehicle tracking on real world surveillance video networks. Such networks consist of heterogeneous cameras with disjoint fields of view and adversarial conditions, including different vehicle orientations, motion blur, occlusion by other objects, multiple scales, and multiple resolutions. These technical challenges fall into two categories:

\squishlist
\item \textbf{Inter-class similarity}: Most vehicles have very similar looks. Vehicles of the same make/type/color appear visually similar due to their manufacturing process. 
\item \textbf{Intra-class variability}: The same vehicle may appear visually different due to different orientations or occlusion. For example, consider the static image recognition approach that trains a classifier through front and rear views of a vehicle, which are very different typically. Since a frame may capture a vehicle at any angle, re-id needs to follow features across orientations. 
\squishend

%\cite{RN8}

\PP{Intra-Class Variability} We focus primarily on the intra-class variability challenge in real world camera networks, ranging from hundreds of cameras in universities to millions in large cities such as London or Beijing. 

One source of variability are the cameras, which are optimized for maximizing coverage under limited numbers. Due to the different purposes of cameras and organic growth of real-world camera networks, they have varied resolutions and capabilities. As a concrete example of critical functions, license plate recognition (LPR) was considered a major consumer of cloud resources in 2017 \cite{RN42}; just a few years later, many 2019 models of video cameras offer LPR as an internal feature, removing the cloud cost. The second source of variability consists of vehicles themselves. The images of vehicles may be partially occluded by objects in the environment, including other vehicles. This problem has been the target of several previous studies \cite{RN20, RN32, RN41} on addressing the intra-class variability problem under testbed assumptions (little variability from cameras). 

Addressing vehicle variability requires global features differentiate shape, color, or brand. Addressing camera variability requires local features from vehicle parts across orientations such as headlight, bumper, or decals.

% Solving the intra-class variability challenge requires extraction of fine-grained features that allow invariant tracking; for example \cite{RN7} tracks 20 hand-engineered key points on vehicles to perform re-id. 
%Typical approaches to re-id consist of CNNs that project images of vehicles into a latent space embedding. The goal is to make embeddings of the same vehicle cluster together, and embeddings of different vehicles further apart. 

\PP{\model} This paper introduces \model, a vehicle re-id model capable of handling both sources of variability (cameras and vehicles) with multiple orientations and conditions. \model, or \textbf{G}lobal and \textbf{L}ocal \textbf{A}ttention \textbf{Mo}dules for \textbf{R}e-ID, integrates global and local attention modules to ensure simultaneous global and local feature extraction. \model's global and local attention allows us to perform re-id with fewer assumptions: we show superior performance in adversarial conditions. Since we extract rich features from global and local features in the same network, our approach is compatible with the inter-class similarity problem: global attention tracks features across vehicles (\eg color) while the local attention differentiates visually similar but distinct identities using local features.

\PP{Contributions} Our technical contributions are:
\squishlist
\item \textbf{Loss Combination \& Normalization.} We use both metric \& softmax loss, and we use normalization to project features from metric loss to softmax loss. Different from work in \cite{RN43}, we examine normalization strategies that are better for re-id than batch norm.

\item \textbf{Global and Local Attention.} We improve global feature extraction by using a global attention network to reduce sparsity of the input conv\footnote{convolutional} layer. Simultaneously, we use a novel local attention mechanism to automatically detect and extract part-based features from the global features. Global \& local features are combined in a unified network. \model learns to extract local features without guidance on vehicle parts \cite{RN81, RN7, RN104}, regions \cite{RN2}, orientations \cite{RN7, RN3, RN4}, or views \cite{RN19, RN36}. \model uses a single unified model compared to multiple branches in \cite{RN19, RN37, RN2, RN8, RN7, RN48, RN36, RN3, RN10, RN4}, making it smaller and faster than current approaches (25\% better mAP, 10x smaller than \cite{RN19}; see \cref{tab:sizes})
\squishend

%Different from \cite{RN81, RN19, RN2, RN7, RN36, RN3, RN4}, 

We evaluate \model on VeRi-776 \cite{RN1}, which contains vehicles with various orientations, and achieve 80.34 and rank-1 96.53\%. On VRIC \cite{RN37}, which contains multi-resolution, multi-scale images with adversarial conditions, we achieve mAP 76.48 and rank-1 78.58\%. On the larger VeRi-WILD \cite{RN38}, we also achieve state-of-the-art results, with mAP 77.15 and rank-1 92.13\%. 
\section{Related Work}
\label{sec:related}
\subsection{Single Camera Tracking}
\label{sec:sct}
Recent vehicle detection and tracking approaches track by clustering detections across multiple frames. Single-camera tracking is framed as a tracklet-based graph model, where vehicle tracklets are clustered by minimizing a cost function \cite{RN86, RN87, RN83, RN85, RN82, RN32}. Features are either CNN-based \cite{RN83, RN82, RN32, RN89, RN88} or histogram-based \cite{RN90, RN91, RN32}. More recently, CNN and histogram features have been combined for more robust single-camera and overlapping-camera tracking in \cite{RN32}. Due to the continuity assumption, single-camera methods are better at combining shorter-trajectory tracklets without large gaps \cite{RN92}.

%where there may be occlusion or unreliable detections between frames

\subsection{Intra-Class Variability}
\label{sec:icv}
More recently, the focus has shifted to addressing the intra-class variability condition that is common in multi-camera setting. To address intra-class variability, a feature extractor should have similar features of the same vehicle regardless of camera source, vehicle orientation, environmental occlusion, detection scale, or frame resolution. The usual approach is to combine global and local feature extraction \cite{RN81, RN19, RN2, RN8, RN9, RN7, RN48, RN3, RN4}. A global feature extractor CNN is paired with supervised local feature extractors. Global and local features are combined with dense layers to create combined features for re-identification. Intra-class variability of features is usually minimized with metric learning, such as triplet loss.

\PP{Datasets} Liu \etal released VeRi-776 \cite{RN1}, with 776 unique vehicle identities, each with images from multiple cameras in different ambient conditions. VeRi-776 has splits of 576 identities for training and 200 identities for evaluation. During evaluation, a single image from each of the 200 identities is used as a query, and the remaining images, called the gallery set, are used for re-id. Performance is evaluated with the mAP and rank-1 metric. More recently, Lou \etal released VeRi-Wild \cite{RN38}, which has significantly more vehicles captured over a larger area (200 sq.km. for VeRi-Wild compared to 1 sq.km. for VeRi-776), with more variability in images. VeRi-Wild’s test set contains 3000 identities, with evaluation similar to VeRi-776. Kanaci \etal released the more realistic VRIC \cite{RN37}, with adversarial conditions such as occlusion, multi-scale, and multi-resolution images (2811 identities in test set).

\subsection{Re-ID with Supervised Local Features}
\label{sec:supervisedreid}
Common approaches for vehicle re-id have used supervision over local features to learn discriminative feature extraction. Wang \etal develop an orientation-invariant approach \cite{RN7} that uses 20 keypoint regions on vehicles to identify common features across orientations. A subnetwork clusters the vehicles based on orientation to extract orientation specific features. All features from keypoints, orientation-specific network, and global feature extraction network (a CNN-based feature extractor) are combined with dense layers to create final features (mAP 51.42 on VeRi-776). Shen \etal  proposed Siamese networks with path proposals \cite{RN6} for multi-camera re-id. Contrastive loss is used on the siamese network to learn global features. An LSTM recurrent network is used to determine path proposal validity to further improve re-identification accuracy by extracting local features of vehicle paths. Features from Siamese and LSTM networks are combined to create final discriminative features (mAP 58.27). Liu \etal propose a region-aware network \cite{RN2} that has sub-models that each focus on a different region of the vehicle’s image, since each region of the vehicle has different types of relevant features. Global and local features are concatenated to generate globally invariant features for re-id (mAP 61.5). Each approach uses triplet loss as the distance metric learning loss for training.

\PP{Triplet Loss} Under triplet loss, training uses three images:  anchor $a$, positive $p$, and negative $n$, where the $x_a$, $x_p$ are images of the same identity and $x_n$ is a different identity. The model $f$ extracts embeddings $e=f(x)$ of each $x$. Triplet loss addresses the constraint $d(e_a, e_p)+\alpha\leq d(e_a, e_n)$ where $d$ is a distance function and $\alpha$ is the margin (see \cref{sec:triplet}); during training, $e_a$, $e_p$ are clustered together while $e_a$, $e_n$ are pushed apart. Effective training requires hard triplets where $d(e_a, e_n)<d(e_a, e_p)$ (here, $e_n$ is a hard negative); in easy triplets, a re-id model cannot learn since triplet loss is 0. Processing images from the entire training set for hard triplets is expensive ($O(n^3)$ runtime \cite{RN13}).

%\subsection{Re-ID with Generative Models}
\PP{GANs in Re-ID} Recently, there has been interest in using generative networks (GAN) \cite{RN44} to create synthetic hard negatives. Generative models are proposed in \cite{RN8, RN9, RN36, RN3, RN10} to address limitations of triplet loss. The GAN's generator creates a synthetic reconstruction image $x'$ from input $x$. The discriminator distinguishes between $x$ and $x'$, acting as a re-id model that can handle hard negatives; typically $x'$ is a hard negative of $x$ since a GAN makes mistakes in reconstruction. A cross-view GAN is proposed in \cite{RN36} to create hard negatives of different orientations to augment easy triplets (mAP of 24.65 on VeRi-776). Zhou \etal propose a view-point aware network \cite{RN3} that combines features from viewpoint feature extractors with a GAN to create cross-view features for re-id (mAP of 61.32). Lou \etal \cite{RN8} use a GAN to generate hard negative cross-view and same-view images; the discriminator of the GAN is used for re-id (mAP of 57.44). Peng \etal \cite{RN9} use a dual branch GAN to generate variations of hard negatives by permuting combinations of image attributes (lighting, vehicle type, background). Style transfer converts between target \& source datasets to augment training data (mAP of 51.01). Zhou \etal \cite{RN10} suggest a multi-view GAN with an LSTM for temporal modeling of paths (mAP of 24.92).

\section{System Overview}
\label{sec:system}
\begin{figure}[t]
	\centering
	\includegraphics[width=3.4in]{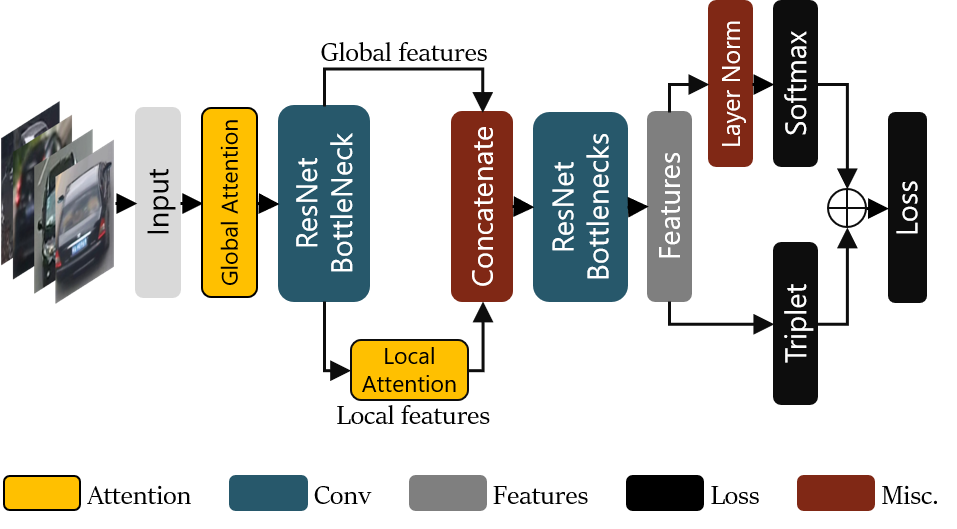}
	\caption{\textbf{\model}: Global Attention reduces input layer sparsity, see \cref{sec:sparsity}. At the first ResNet  bottleneck we apply Local Attention to automatically extract local part-based features. Local features are concatenated with global features and passed through remaining bottlenecks in ResNet to extract rich features for re-id}
	\label{fig:system}
\end{figure}

We now describe \model, our end-to-end re-id model with global and local attention for self-guided local feature extraction, shown in \cref{fig:system}. We will first present a simple and strong base model for vehicle re-id that outperforms several current approaches (see \cref{sec:baseeval}). We then cover our proposed improvements: (i) loss combination \& normalization, and (ii) global \& local attention to ensure global feature plus self-guided part-based local feature extraction.

\subsection{Base Model}
\label{sec:baseline}
% Our base model uses 10x fewer parameters (approx. 11M) to achieve mAP of 64.4 on VeRi-776, compared to 110M parameters and 64.6 mAP in \cite{RN19}, see \cref{tab:sizes}.
% feature \textit{extraction} and feature \textit{interpretation}

A vehicle re-id model performs two tasks: (i) feature \textit{extraction} identifies relevant global and/or local features from images using conv layers, and (ii) feature \textit{interpretation} projects these conv features into a latent space for metric learning. Adding dense layers in feature extraction forces the dense layers to perform feature interpretation, while the conv layers perform feature extraction \cite{RN12}. This approach is common in existing multi-stage, multi-network re-id \cite{RN2, RN8, RN7, RN3}. Since conv layers focus on pixels with a spatial constraint, using them for feature interpretation is more effective in tracking image features \cite{RN902,RN903}. %Further, by learning feature combinations for classification, dense layers overfit on the combinations in the training set and cannot classify novel combinations in the re-id test sets \cite{RN901}.

For our base model, we remove all dense layers and use the output of the ResNet backbone as the final features for re-identification. Removing dense layers forces the convolutional backbone to learn both feature extraction and feature interpretation simultaneously. Relying on conv layers to perform interpretation is better than relying on dense layers, since conv layers capture spatial constraints.

So, our base model is a ResNet with stride of 1 in the final pooling layer\footnote{stride 1 improves feature extraction and reduces information loss \cite{RN46}} and no dense layers for the output features. We use two tricks recently proposed in \cite{RN43}: random erasing and warmup learning rate. The base model uses only global features for re-id and achieves mAP competitive with or better than existing works that use supervisory sub-models such as \cite{RN7,RN4,RN2,RN81}, see \cref{sec:baseeval}. Our advantage lies in removing the dense layers, indicating the global features are sufficient for reasonable performance. 

% As described in related work have opted for global and local feature extraction sub-networks. Multiple sub networks are trained with supervision to perform fine-grained feature extraction. Global and local fine-grained features are combined using dense layers. 

\PP{Random Erasing Augmentation} We use random erasing augmentation from \cite{RN15} to improve fine-grained feature extraction for re-id. Occlusion is simulated by randomly erasing a rectangular region within the image with probability $p_r$, where we set $p_r=0.5$. %Erased pixels are replaced with the mean of the image. The erased rectangle dimensions $H_e,W_e$ are obtained by $r_e H_I,r_e W_I$ where $H_I,W_I$ are height and width of the image; we set $0.01\leq r_e \leq 0.5$.

\PP{Warm-up Learning} We also use the warm-up learning rate from \cite{RN43}. For a given base learning rate $l_r$, we start with $0.1l_r$ at epoch 0 and linearly increase to $l_r$ by epoch 10. We test two versions of the warmup learning rate: warmup-1, where we linearly increase the learning rate every epoch, and warmup-2, where we increase every 2 epochs.

\subsection{\model: Improvements on Base Model}
\label{sec:improvements}
\model incorporates the modifications described in \cref{sec:lossnormalization} and \cref{sec:iaandta} to the base model to deliver superior re-id performance. We briefly introduce the improvements:

% Several approaches propose combining multiple loss types to improve fine-grained feature extraction for the intra-class variability problem; a variation of triplet loss is proposed in \cite{RN20}, a quadruplet loss is used in \cite{RN4}, and combinations of feature losses are used in \cite{RN48}. : the triplet loss performs coarse-grained clustering, and the softmax loss improves fine-grained feature extraction

\squishlist
\item \textbf{Softmax Loss \& Normalization.} We use triplet loss with softmax loss, similar to person re-id \cite{RN43}. Since triplet loss minimizes $l_2$ norm while softmax minimizes intra-class cosine distance, we use normalization between triplet and softmax features. Differently from \cite{RN43}, we will show that layer and group norm yield better results since they do not rely on batch size.

\item \textbf{Global and Local Attention} We develop a \textbf{global} attention module that regularizes global feature maps after the first convolutional layer to reduce the activation sparsity. This allows more global features for the feature extractor, increasing feature diversity. We then develop a novel \textbf{local} attention module to perform self-guided local feature extraction. Local attention allows the re-id model to learn important part-based features during training with self-guidance, compared with supervised local feature extractors in \cite{RN81, RN7, RN48}. The unsupervised feature maps generated by local attention correspond to part-based features such as headlights, emblem, and license plate (see activations in \cref{fig:partfeatures}). Global and local features are combined in a unified network for a small, fast, and superior re-id.
\squishend

Together, these improvements create \model, which uses global and local attention to deliver superior re-id performance compared to the state-of-the-art on VeRi-776 \cite{RN1}, VRIC \cite{RN37}, and VeRi-Wild \cite{RN38}, see \cref{sec:evaluation} and \cref{tab:evaluation}.
\section{Softmax Loss and Normalization}
\label{sec:lossnormalization}
Recent methods for person re-id have used both metric and classification loss. So, we combine triplet with softmax loss and normalize triplet loss features for softmax loss.

\subsection{Standard Triplet Loss}
\label{sec:triplet}
The standard triplet loss is useful in the re-id scenario to learn feature separation in $l_2$ space so that identities can be clustered using component features (compared to softmax loss, metric learning is extensible to feature combinations not seen in training \cite{RN16}). The triplet loss is formulated as:

\begin{myequ}
\label{eq:triplet}
L_{Triplet} = \Sigma \parallel a-p \parallel_2^2 - \parallel a-n \parallel_2^2 + \alpha
\end{myequ}

where $a$,$p$,$n$ are the anchor, positive, and negative of a triplet (see  \cref{sec:supervisedreid}) and $\alpha$ is the margin constraint. $\alpha$  enforces the minimum distance difference between two images from the same identity versus images from distinct identities. With triplet loss, unique identities are mapped to the same or nearby point(s). $\alpha$ ensures clustering of similar identities and dispersion of dissimilar identities. 

\subsection{Batched Hard Mining}
\label{sec:batchhardmining}
Effective training under triplet loss requires appropriate sampling to ensure hard negatives. As discussed in \cref{sec:supervisedreid}, the choice of triplets can improve convergence and accuracy. Using all triplets saturates training with `easy' cases where $a$, $n$ can be discriminated on coarse features alone, \eg color, type, or model. Re-id models trained with easy triplets learn only coarse features without details \cite{RN16}.

In hard negative triplets where $d(e_a, e_n)<d(e_a,e_p)$, the network must learn to push the negative further away and the positive closer to the anchor. Finding hard triplets by searching the training set is an $O(n^3)$ search \cite{RN13}. The batch hard strategy \cite{RN16} proposes performing this search in the mini-batch setting. We apply batch hard mining and use only the hard negatives for training the re-id model.

% With a large enough mini-batch (\eg 64 images of 16 ids, 4 images per id), there will some hard triplets. 

\subsection{Softmax Loss}
Softmax loss projects features around the origin and separates classes by maximizing cosine distance between class features around the hypersphere:

\begin{myequ}
\label{eq:softmax}
L_{softmax} = \Sigma_i \shortminus q_i\log p_i\,\,\,\,\,\,\,\,\,\,\,q_i=\mathds{I}(\hat{y}=i)
\end{myequ}

We use the softmax loss with label smoothing \cite{RN18}, where the true zero values are replaced with $\epsilon$ to reduce classifier over-confidence; we let $\epsilon=1/N$, where N is the number of identities in the training set, so

\begin{myequ}
\label{eq:softsmooth}
q_i=\mathds{I}(\hat{y}=i)-(1/N \sgn(\mathds{I}(\hat{y}=i)-0.5))
\end{myequ}

The smoothed softmax loss is effective in reducing overfitting on feature combinations in classification \cite{RN18}. 

\subsection{Combining Triplet and Softmax Loss}
Softmax loss carries higher discriminative ability since it does not need to generalize to unseen classes. Triplet and softmax losses work towards different targets. Triplet loss, by minimizing intra-class and maximizing inter-class $l_2$ norm, works on each feature independently. Softmax loss constraints the feature space to the hypersphere around the origin and learns separation only for training classes. Novel feature combinations in the testing set cannot be projected in a model trained with softmax loss only \cite{RN16}.

Combining triplet and softmax losses offers a compromise between the two. Triplet loss aids in generalization to unseen feature combinations by dealing with each feature independently. Simultaneously, softmax loss improves fine-grained features. The final loss for our model is:

\begin{myequ}
\label{eq:losses}
L_{TriSoft}= L_{Triplet} + L_{Softmax}
\end{myequ}

\subsection{Feature Normalization}
We now cover feature normalization to separate triplet loss features from softmax loss features. We perform this due to the differences in loss calculation: the triplet loss calculates distance between individual embeddings and minimizes intra-class $l_2$ norm, while the softmax loss minimizes intra-class cosine distance around the unit hypersphere. We first describe four common normalizations below:

%We propose using layer and group normalization, differently from batch normalization proposed in \cite{RN43}.

%Then we explore their impact on the triplet features and describe our choice of group and layer normalization.

\PP{L2 Normalization} L2 normalization rescales features s.t. $\sum f_i^2=1$. It has been used to regularize the output features of metric learning losses and remove feature skew in person re-id \cite{RN94, RN95} and face recognition \cite{RN13}.

\PP{Batch Normalization} Each of batch, group, \& layer norm perform normalization on a given feature set $x_S$, where normalized feature $x'_S$ is obtained by $x'_S=(x_S-\mu_S )/\sigma_S$. In batch normalization, the feature set $x_S$ is the set of all pixels sharing the same channel index. Normalization is performed across height, width, and batch axes.

\PP{Group Normalization} Group normalization \cite{RN93} sets $x_S$ as the pixels in the same batch that are co-located channel-wise \cite{RN93}. Given a grouping constant $G$, $x_S$ is normalized across height \& width, along with contiguous sets of $C/G$ channels, where $C$ is the total number of channels.

\PP{Layer Normalization} Layer normalization \cite{RN96} is a special case of group normalization when $G=1$. $x_S$ is the set of pixels in the same batch, and normalization occurs across the entire batch on height, width, \& channel axes.

\subsection{Impact of Each Normalization}
\label{sec:impact}

\begin{figure}[t]
	\centering
	\includegraphics[width=3.4in]{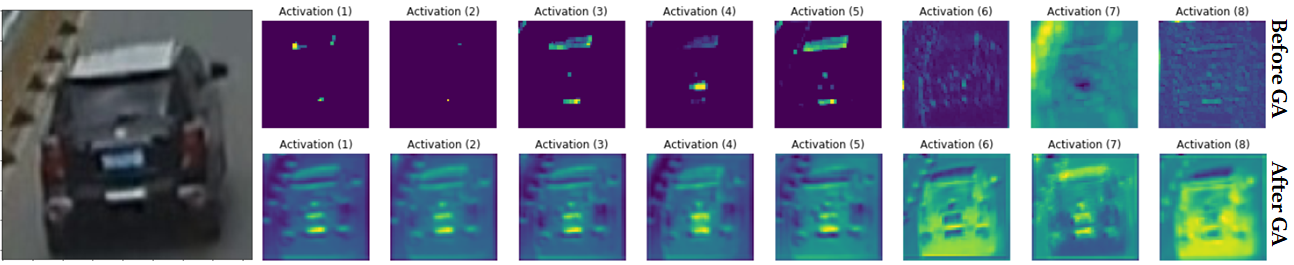}
	\caption{\textbf{Activation Regularization}: Activation correction after global attention. Images 1, 5, 6 are skewed towards dark shadow, which does not carry any information. Global attention reduces sparsity  (see \cref{sec:sparsity}) by allowing more features to the backbone.}
	\label{fig:iafirst}
\end{figure}

%Softmax loss discriminates between individual classes by building $p$-dimensional hyperplanes, where $p$ is the number of identities in the training set. Loss minimization involves moving embeddings away from the hyperplanes to enforce discrimination and maintaining inter-class separation by maximizing cosine distance between classes. 

In our case, $l_2$ normalization fails since we require projection from triplet loss features to softmax features, which are located around the unit hypersphere. Since $l_2$ normalization only rescales features, it is ineffective.

Batch normalization is proposed in \cite{RN43} to decouple the losses and project the triplet loss around the unit hypersphere to create features more amenable to softmax loss. Batch normalization is sensitive to the batch size \cite{RN93} and has reduced impact as training progresses. Since we use only hard triplets for training (\cref{sec:batchhardmining}), the true batch size changes during training, reducing batch norm efficacy.

So, we use group and layer norm to adapt to the dynamic batch size from batch hard mining. Since neither rely on the batch size, they are attractive substitutes for batch norm. We use layer norm to convert triplet loss features to softmax features. Layer norm is preferred over group norm since it assumes equal contribution of all channels \cite{RN96}. In addition to layer norm for decoupling triplet and softmax loss, we replace batch norm inside ResNet with group norm. Our choice of group over layer norm here is based on \cite{RN96} and \cite{RN93}, where the authors find the layer norm assumption of equal channel contribution is less valid for conv layers. 	
\section{Global Attention and Local Attention}
\label{sec:iaandta}
We now cover our global attention and local attention modules. Our goal with global attention is to improve global feature extraction and ensure richer features for re-id. In conjunction, local attention allows our model to track part-based features across cameras; while part based features have been used for re-id, our novelty is in self-guided local attention, where part-based features are learned without guidance. This allows our model flexibility in selecting which parts are relevant in a vehicle image.

\subsection{Global Attention (GA)}
\label{sec:ia}

\begin{figure}[t]
	\centering
	\includegraphics[width=3.4in]{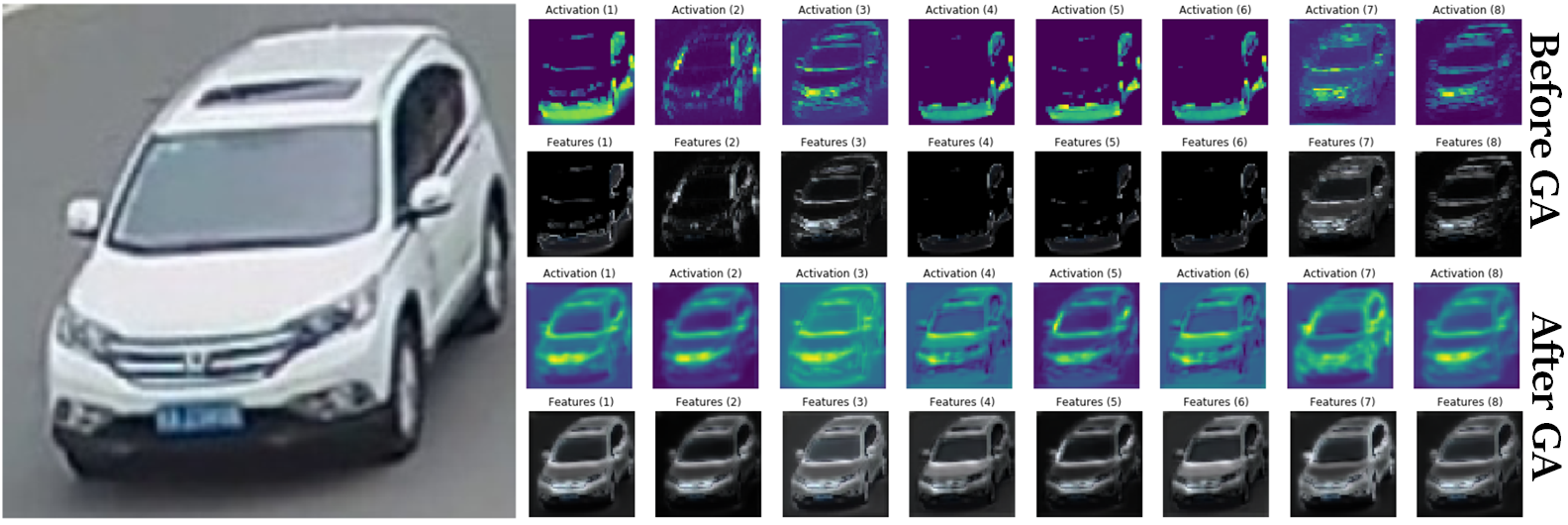}
	\caption{\textbf{GA Features and Activations}: Most feature maps carry no information before GA (first and third rows). Addition of GA improves information transfer to the remaining layers. By reducing sparsity, deeper layers get access to more global features.}
	\label{fig:iasecond}
\end{figure}

We first develop an attention-based regularizer called global attention to fix activations in the first conv layer. The first conv layer is crucial in feature extraction since its information is propagated through the entire network. We find during training the base model from \cref{sec:baseline} that the activations in the first layer are sparse or skew towards irrelevant features like shadow (\cref{fig:iafirst}, upper row). We propose the global attention (GA) module to reduce sparsity and improve feature extraction (\cref{fig:iafirst}, lower row). 

The GA module is inserted after the first conv layer of ResNet before the basic blocks. The attention procedure uses two $3\times 3$ convolutional layers with Leaky ReLU activation to retain negative weights from the first convolutional layer. The output is passed through a sigmoid activation and element-wise multiplied with the input: 

\begin{myequ}
\label{eq:iamodule}
GA(F)=F\cdot \mathnormal{Sig}(f^{3\times 3} (\mathnormal{LeakyReLU}(f^{3\times 3} (F))))
\end{myequ}

In contrast to CBAM \cite{RN11}, which learns spatial and channel attention independently and applies the same spatial map to all channels, we learn feature correction for each kernel. We also do not use any pooling, since both max and average pooling cause loss of information between features.

As we see in \cref{fig:iafirst}  and \cref{fig:iasecond}, before GA there are several feature maps that do not encode any information. Applying GA corrects the feature maps by allowing more information through the network. GA is only useful with the first convolutional layer, since it indiscriminately allows features; adding GA in deeper layers would reduce feature discrimination and limit learning by reducing sparsity \cite{RN101,RN103}. In the first layer, however, it acts as a regularizer by reducing layer sparsity and ensuring more information from the image is passed into the re-id model (see \cref{sec:sparsity}).

\subsection{Local Attention (LA)}
\label{sec:ta}

\begin{figure}[t]
	\centering
	\includegraphics[width=3.4in]{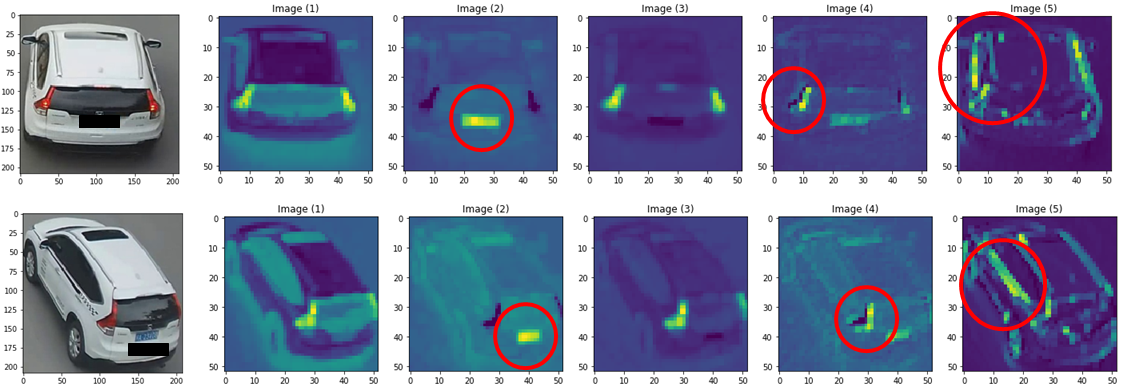}
	\caption{\textbf{Local Attention}: Local attention automatically detects part-based features for vehicles without supervision on parts. These self-guided local features correspond to actual part-based features. Channels 1,3 track both rear lights. Channel 2 tracks license plate only. Channel 4 tracks left rear light. Channel 5 tracks the dark decal below the vehicle windows. Best viewed in color.}
	\label{fig:partfeatures}
\end{figure}

Part-based features allow models to focus on common vehicle parts independent of global features. This augments softmax loss, which improves fine-grained feature extraction on the entire image. Part-based feature extraction is a common aspect of several recent works in person and vehicle re-id \cite{RN81, RN2, RN46, RN85, RN7, RN48, RN3}. Intuitively, using both global and local part-based features allows a model to learn globally invariant details, such as vehicle type (\eg sedan, truck, SUV) and color, as well as locally invariant details from common parts, \eg headlights or emblem.

Current methods discussed in \cref{sec:supervisedreid} use supervised local features, where subnetworks are trained to detect specific vehicle parts and extract features from these parts independent of the global feature extractor backbone. The model in \cite{RN7} uses supervision to detect 20 keypoints on vehicles, \eg left headlights and right headlights. Other approaches use supervision to select region \cite{RN2} or orientation features \cite{RN3}. Finally, GAN-based approaches \cite{RN8, RN9, RN10} use synthetic negatives as a supervising signal for local features.

We propose Local Attention (LA) for \textit{unsupervised} part-based feature extraction. We apply spatial and channel attention \cite{RN11} at the first ResNet bottleneck only  to impose attention on the feature maps (see \cref{fig:system}), allowing our model to learn discriminative local regions for each vehicle image and extract these local features in conjunction with global features from the entire image. In contrast to the soft attention mechanism discussed in \cite{RN7}, convolutional attention learns weights for different regions of the image, as opposed to weights for each channel of the convolutional features. LA deeper layers would reduce the effectiveness of part-based feature extraction since feature map dimensions are already small relative to the input image in re-id: we use input image size of $208\times 208$ pixels, and the output of the second basic block has images of size $27\times27$; most of the spatial information has been moved to the channels.

\PP{Local Feature Extraction} After obtaining the global features $F_G$ from the first ResNet basic block after GA, we apply spatial and channel attention to obtain local features $F_L$, where $F_G,F_L\in \mathds{R}^{H\times W\times C}$ with the same dimensions. We then sum the global and local features after applying a channel-wise mask, so that half of our final features come from $F_G$ and half of our features from $F_L$: 

\begin{myequ}
\label{eq:masking}
F=M_G \odot F_G + M_L \odot F_L
\end{myequ}

where $M_G,M_L\in \mathds{R}^C$, $M_L=\bar{M_G}$, and for each $m_i\in M_G$, $m_i=0\,\,\forall i<\lfloor C/2\rfloor$ and $m_i=1\,\,\forall i\geq \lfloor C/2 \rfloor$.

Since we let the local attention learn attention regions of the image during training, we do not require supervision to detect keypoints: we show in \cref{fig:partfeatures} that the local features learned by local attention module correspond to real part-based features including rear lights, license plate, and side decals for two images of the same vehicle.

\begin{figure}[t]
	\centering
	\includegraphics[width=3.4in]{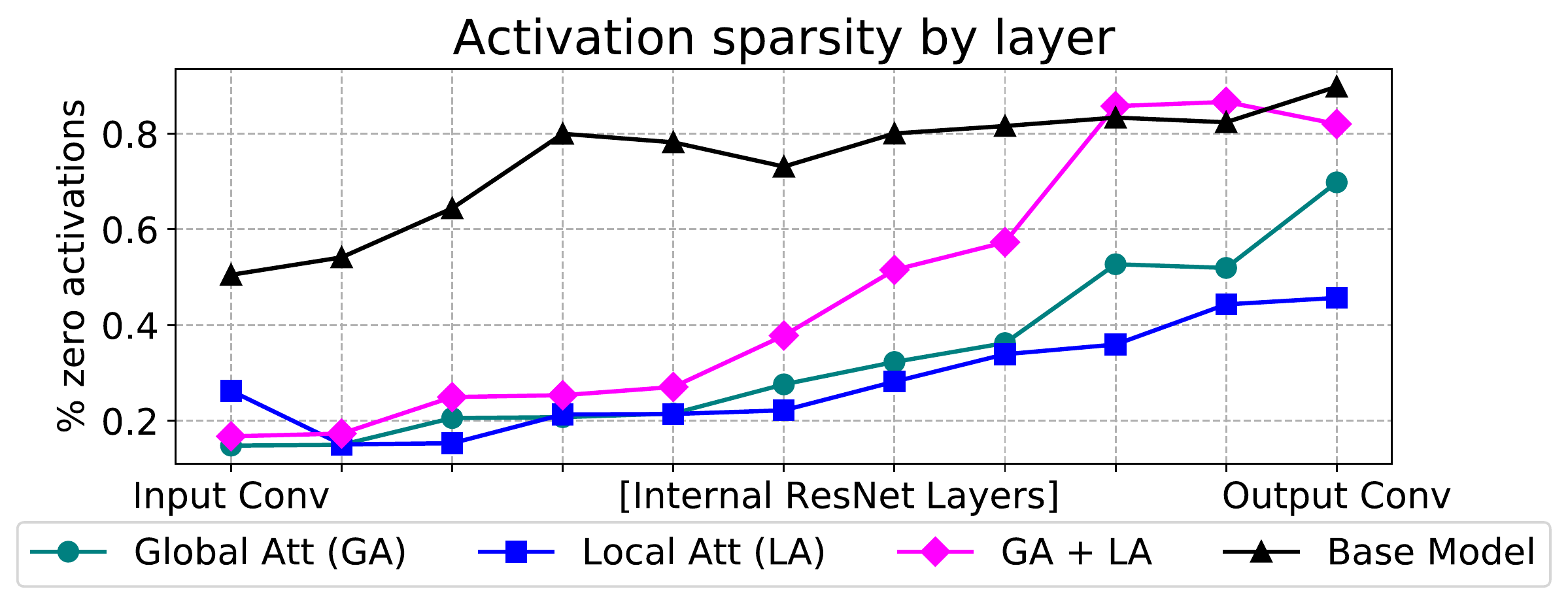}
	\caption{\textbf{Layer Sparsity}: Sparsity of activations by layer for input convolution, output feature, and ResNet's internal conv layers. The base model maintains high sparsity throughout, while LA retains low sparsity. GA ensures low sparsity for input conv and highs sparsity for output conv.}
	\label{fig:sparsity}
\end{figure}

\subsection{Model Sparsity with Attention}
\label{sec:sparsity}

\begin{figure*}[t]
	\centering
	\includegraphics[width=6.8in]{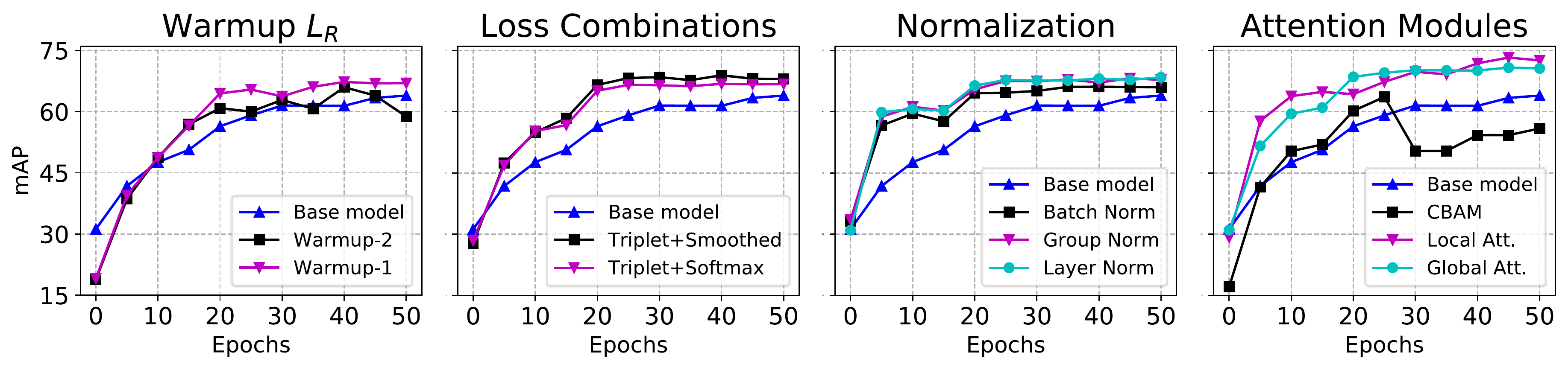}
	\caption{\textbf{Evaluation}: Impact of warmup, losses, normalization, \& attentions. Y-scale is same for all plots. Each model is evaluated on VeRi-776; images of the same ID from same camera are removed from gallery set \cite{RN1}. See \cref{sec:baseeval} \& \cref{sec:evalimprove} for mAP values and discussion.}
	\label{fig:evaluations}
\end{figure*}

% . Warmup-1 is better than warmup-2 since it increases $l_r$ slower. Using both triplet and softmax loss improves ranking performance. The label-smoothed softmax loss performs slightly better than the non-smoothed softmax loss. Normalization similarly improves performance, with layer norm outperforming batch norm because of batch size agnosticy. Global \& local attention outperform both the base model and CBAM applied to all layers (GA: 71.08, LA: 73.28, CBAM: 63.59 mAPs).

An important consideration in deep networks is weight sparsity. Sparse network activations, where only a subset of a feature extractor network’s weights are activated, are desired since they allow compressible and fast networks that can perform less calculations than non-sparse counterparts. While neural networks can tolerate high levels of sparsity \cite{RN100, RN101}, it is less desirable in the first and last layers since these are disproportionately important for feature extraction \cite{RN12, RN102, RN103, RN100}. As we noted in \cref{sec:ia}, global attention reduces the sparsity of the first conv layer (see \cref{fig:sparsity}).

We calculate sparsity by measuring activations in the feature kernels from a random subset of images and calculating fraction of zero-valued kernels. The inclusion of the LeakyReLU activation in the GA module allows the re-id model access to more global features. Combining input attention (global) with the targeted attention (local) reduces sparsity of the first conv layers while increasing them for the deeper feature extractor layers. This increased sparsity can reduce computation for real-time networks \cite{RN102, RN103, RN101}. We leave further study of model sparsity \& impact on serving for real-time inference for future work. 

\section{Evaluation}
\label{sec:evaluation}
\subsection{Experimental setup}

\PP{Datasets} We use the VeRi-776 dataset proposed in \cite{RN1} to compare against related work (200  ids in test set), the recently proposed VRIC dataset \cite{RN37} (2811 ids in test set), and the larger VeRi-WILD dataset \cite{RN38} (3000 ids in test set). 

\PP{Minibatch selection} For each of the datasets, the experimental setup remains the same. During training, we use a batch size of 36 samples, with 6 unique ids per batch. 

\PP{Metrics} The pairwise distance matrix between gallery image embeddings and query image embeddings are used to sort gallery images for each query. We use mAP metric to measure overall ranking, as well as the rank-1 score.

\subsection{Base Model}
\label{sec:baseeval}

Our base model uses ResNet18 as the backbone feature extractor, with group norm  instead of batch norm. We show competitive results with our smaller model  compared to most existing works that use larger models as the backbone extractor plus submodels, indicating our base model construction method is robust and well-suited for re-id (see \cref{tab:sizes}). Images are resized to $208\times208$ with (i) random flipping, $h=0.5$, (ii) normalization, and  (iii) random erasing with $p_r=0.5$. We use only triplet loss, with $\alpha=0.3$ and batch hard mining \cite{RN16} and train with Adam optimizer, with $l_r=$1e$\shortminus$4, decaying every 20 epochs with decay factor 0.6. The base model achieves mAP of 64.48, with rank-1 63.9\% and rank-5 86.2\% after training for 100 epochs.

\PP{Warmup Learning Rate} With warmup learning rate \cite{RN43}, we achieve higher mAP. We test warmup-1 and warmup-2 (see \cref{sec:baseline}), with base learning rate 1e$\shortminus$4. Then, $l_r$ is decayed with $\gamma=0.6$ every 20 epochs. Warmup-1 achieves mAP of 67.34; warmup-2 achieves lower mAP of 65.98, indicating smaller increases are more useful for re-id.

\subsection{Proposed Improvements}
\label{sec:evalimprove}

\begin{table*}[t]
	
	\centering
	\caption{\model evaluated on VeRi-776, VRIC, and VeRi-Wild with mAP, Rank-1, and Rank-5 metrics.}
	\begin{tabular}{l|rrr|rrr|rrr}
		\hline
		\multicolumn{1}{c|}{\multirow{2}{*}{\textbf{Approach}}} & \multicolumn{3}{c|}{\textbf{VeRi-776}}           & \multicolumn{3}{c|}{\textbf{VRIC}}               & \multicolumn{3}{c}{\textbf{VeRi-Wild}}          \\
		\multicolumn{1}{c|}{}                                   & mAP            & Rank@1         & Rank@5         & mAP            & Rank@1         & Rank@5         & mAP            & Rank@1         & Rank@5         \\ \hline
		\textbf{GLAMOR (Ours)}                                  & \textbf{80.34} & \textbf{96.53} & {\ul 98.62} & \textbf{76.48} & \textbf{78.58} & \textbf{93.63} & \textbf{77.15} & {\ul 92.13}    & \textbf{97.43} \\ \hline
		PGAN \cite{RN104}                      & 79.30          &  {\ul 96.50} & -              & -              & {\ul 78.00}          & {\ul 93.20}          & 74.10          & \textbf{93.80} & -              \\
		BNN-ReID \cite{RN43}                   & 77.15          & 95.65          & 97.91          & -              & -              & -              & 74.27          & 90.43          & 96.85          \\
		MTML-OSG-RR \cite{RN19}          & 68.30          & 92.00          & 94.20          & -              & -              & -              & -              & -              & -              \\
		%QD-DLF  \cite{RN4}                     & 61.83          & 88.50          & 94.46          & -              & -              & -              & -              & -              & -              \\
		%RAM  \cite{RN2}                        & 61.50          & 88.60          & 94.00          & -              & -              & -              & -              & -              & -              \\
		%VAMI+ST  \cite{RN3}                    & 61.32          & 85.92          & 97.70          & -              & -              & -              & -              & -              & -              \\
		GSTRE  \cite{RN20}                     & 59.47          &  {\ul 96.24} & \textbf{98.97}          & -              & -              & -              & -              & -              & -              \\
		%PATH-LSTM  \cite{RN6}                  & 58.27          & 83.49          & 90.04          & -              & -              & -              & -              & -              & -              \\
		Hard-View-EALN  \cite{RN8}             & 57.44          & 84.39          & 94.05          & -              & -              & -              & -              & -              & -              \\
		FDA-Net  \cite{RN38}                   & 55.49          & 84.27          & 92.43          & -              & -              & -              & 35.11          & 64.03          & 82.80          \\
		OIFE+ST  \cite{RN7}                    & 51.42          & 68.30          & 89.70          & -              & -              & -              & -              & -              & -              \\
		MSVR  \cite{RN37}                      & 49.30          & 88.56          & -              & -              & 46.61          & 65.58          & -              & -              & -              \\
		Scale-224-SS \cite{RN37}                               & 47.37          & 88.37          & -              & -              & 43.62          & 62.77          & -              & -              & -              \\
		%Scale-160-SS \cite{RN37}                               & 46.81          & 87.43          & -              & -              & 43.55          & 61.88          & -              & -              & -              \\
		%DAVR  \cite{RN9}                       & 26.35          & 62.21          & 73.66          & -              & -              & -              & -              & -              & -              \\ 
		\hline
	\end{tabular}
	\label{tab:evaluation}
\end{table*}

We now cover our proposed improvements: (i) loss combination plus normalization from \cref{sec:lossnormalization} and (ii) global plus local attention from \cref{sec:iaandta}; results are shown in \cref{fig:evaluations}.

\PP{Triplet and Softmax Loss}
We compare the base model, which uses only triplet loss, with triplet plus softmax and triplet plus smoothed softmax with smoothing parameter $\epsilon=0.2$. The triplet plus softmax achieves mAP 67.51. Using label smoothed softmax improves mAP to 68.89. 

%Compared to recent work in vehicle re-id that uses only triplet loss of combinations of metric learning losses, our approach in adding a classification loss yields better results. While such an approach has been used in person re-id, it has not yet been used for vehicle re-id. Our results clearly indicate its usefulness in this task. 

\PP{Normalization} We examine three normalization policies between loss features: $l_2$ norm, group norm, and layer norm and  compare to base model and batch norm from \cite{RN43}:

\squishlist
\item \textbf{Batch norm.} Batch norm's impact is covered in \cref{sec:impact}. It improves over the base model, with mAP 66.10.

\item \textbf{L2-Normalization.} Since $l_2$ norm only rescales the triplet features, it fails with the softmax loss layer. Training is unstable and requires hyperparameter tuning for convergence; $l_2$ norm achieves mAP 42.75.

\item \textbf{Group Normalization.} Since group norm operates across the spatial \& channel dimensions, it is agnostic to batch size. We use default group size of 16 channels. Group norm outperforms batch norm due to batch size agnosticy and achieves mAP 67.12.

\item \textbf{Layer Normalization.} Since all channels are normalized together, there are no changes in distribution in contiguous feature groups. This is useful for output features since we need each output feature to contribute equally to the distance metric. Layer norm achieves the best results, with mAP of 68.45.
\squishend

\PP{Attention Modules}
We test our global attention (GA) and local attention (LA) modules. By reducing sparsity of the input conv layer (see \cref{sec:sparsity}), GA allows the backbone access to more global features, aiding in richer feature extraction. LA extracts part-based features in the first bottleneck (see \cref{sec:ta}) and passes them to the remaining feature extractor bottlenecks.  We compare to CBAM \cite{RN11} on all layers, which overfits the re-id model on the training set. Since feature combinations in the testing set are novel in the re-id setting, CBAM alone is not enough in guiding attention to relevant regions of the vehicle image. GA and LA improve over the base model due to their global and local feature extraction, with mAPs 71.08 and 73.28, respectively, compared to mAP 63.59 for CBAM (worse than base model).

%\begin{table}[h]
%	\centering
%
%	\caption{Comparison of attention modules to baseline, CBAM \cite{RN11}}
%	\begin{tabular}{lrrr}
%		\hline
%		\textbf{Model}              & \multicolumn{1}{l}{\textbf{mAP}} & \multicolumn{1}{l}{\textbf{Rank@1}} & \multicolumn{1}{l}{\textbf{Rank@5}} \\ \hline
%		Base                        & 64.48                            & 63.9                                & 86.2                                \\
%		CBAM                        & 63.59                            & 79.2                                & 91.06                               \\
%		\textbf{Input Attention}    & \textbf{71.08}                   & \textbf{89.21}                      & \textbf{95.47}                      \\
%		\textbf{Targeted Attention} & \textbf{73.28}                   & \textbf{91.66}                      & \textbf{96.48}                      \\ \hline
%	\end{tabular}
%	\label{tab:evaluations}
%\end{table}

\subsection{Evaluation of \model}

We build our combined Global and Local Attention Model for ReID (\model) with the following components: ResNet18 architecture core, warmup-1 policy, triplet with label smoothed softmax loss, layer normalization, global attention, and local attention. Results on VeRi-776, VRIC, and VeRi-Wild are provided in \cref{tab:evaluation}.

\PP{VeRi-776 (mAP 80.34, R-1 96.53\%)} Performance of \model is similar to PGAN \cite{RN104}, which uses a guided local feature extractor. \model's advantage over PGAN is our self-guided learning with local attention combined with global attention. \model automatically detects local features, while PGAN requires training an object detector. Finally, \model uses ResNet-18 as the backbone, while PGAN uses ResNet50 plus additional submodels.

\PP{VRIC (mAP 76.48, R-1 78.58\%)} \model performs well on VRIC, which has more adversarial conditions. Compared to \cite{RN37}, we use a single model for different scales \& resolutions and achieve higher mAP \& rank-1. 

\PP{VeRi-Wild (mAP 77.15, R-1 92.13\%)} \model also performs well on VeRi-Wild, which is larger than VeRi-776 or VRIC. While our rank-1 is slightly lower than \cite{RN104}, mAP is higher, indicating better overall ranking \&  robustness. Our model is smaller, and furthermore, it is self-guided, compared to supervised local features in \cite{RN104}.

\subsection{\model Size comparison} 
\model extracts re-id features on a single feature extractor backbone by combining global and local features. We show in \cref{tab:sizes} the approximate number of parameters in current approaches and \model plus mAP on VeRi-776. By leveraging global attention to increase number of global features while exploiting part-based feature extraction with local attention, we achieve 25\% better mAP than \cite{RN19} with an order of magnitude fewer parameters.

\begin{table}[h]
	\small
	\centering
	\caption{Size comparison between \model and related works}
\begin{tabular}{lrrr}
	\hline
	\textbf{Approach}                     & \textbf{Params (Est.)} & \textbf{mAP} & \textbf{Rank-1}   \\ \hline
	OIFE \cite{RN7}      & 521M            & 51.42       &   68.30\\
	VAMI \cite{RN3}                                  & 280M            & 61.32       &   85.92\\
	RAM \cite{RN2}       & 164M            & 61.50        &  61.50\\
	QD-DLF \cite{RN4}                                & 40M             & 61.83         & 88.50 \\
	MTML-OSG \cite{RN19} & 110M            & 64.60       &  92.00 \\ \hline
	\textbf{\model (Ours)}                & \textbf{11M}    & \textbf{80.34} & \textbf{96.53}\\ \hline
\end{tabular}
	\label{tab:sizes}
\end{table}

\section{Conclusions}
\label{sec:conclusion}
We have presented \model, a small and fast model for vehicle re-id that achieves state-of-the-art results on a variety of re-id datasets. \model extracts additional global features and performs self-guided local feature extraction using global and local attention, respectively. By using global features to track invariants across vehicles (\eg color) and local features to track invariants across cameras (\eg brakelights, emblem, decals), \model addresses intra-class variability while maintaining compatibility with the inter-class similarity challenge discussed in \cref{sec:intro}. \model outperforms current approaches in re-id while being $\sim$10x smaller. \model also achieves state-of-the-art results without supervision on local features, thereby removing bottlenecks due to generating training data for supervised part-based features.

{\small
\bibliographystyle{ieee_fullname}
\bibliography{main}
}

\end{document}